# Formulation of probability theory problem with subtle condition

Rafayel Petrosyan[1,2]

1. Zaven & Sonia Akian College of Science and Engineering, American University of Armenia, Yerevan 0019, Armenia

2. L. A. Orbeli Institute of Physiology, National Academy of Sciences of the Republic of Armenia, Yerevan 0028, Armenia

**Abstract**

Problems in probability theory prove to be one of the most challenging for students. Here, we formulate and discuss four related problems in probability theory that proved difficult for first to fourth-year undergraduate students whose first language was not English. These examples emphasize how crucial it is to understand the conditions and requirements of the problems precisely before starting to solve them. We discuss the solutions to those problems in detail, complement them with numerical estimations, and link the conditions in the problems to the logical statements in Python programming language. We also tested two widely used chatbots (GPT-4o and Claude 3.5 Sonnet) by checking their responses to these problems.

**Introduction**

Probability theory is essential in many fields, including physics, engineering, data science, finance, and psychology. Hence, it was central to our "Mathematical Thinking" course taught at the American University of Armenia (AUA). First to fourth-year undergraduate students pursuing Bachelor of Arts degrees in Business, English & Communications, and Politics and Governance took this course whose first language was not English.

There are problems in probability theory that are counterintuitive but have definite answers, such as the Monty Hall problem (1), ambiguously formulated where, for example, depending on the sampling procedure, the answers can differ, such as Bertrand paradox and The Two Child problem (2), and still unsolved in the realm of probability theory and philosophy such as Sleeping Beauty problem (3). Here, we formulate and discuss several related problems that have proved challenging for undergraduate students not majoring in STEM. First, we present the problems, then their solutions and the analyses. Next, we computationally verify the answers through simple Python code. Finally, we discuss and classify the responses of two popular chatbots, GPT-4o (4) and Claude 3.5 Sonnet (5), to these problems.

———————————————————

Correspondence to: rafayel.petrosyan@aua.am

OrcidID: https://orcid.org/0000-0003-1888-0379

**Formulation of problems**

Below, we formulate several problems from probability theory about two randomly chosen people and their birthdays:

1. *What is the probability that two randomly chosen people were not born in the same month?*
1'. *What is the probability that two randomly chosen people were not born in May?*
2. *What is the probability that two randomly chosen people were not both born in May?*
   Equivalent to: *What is the probability that at least one of the two randomly chosen people was not born in May?*
3. *What is the probability that neither of the two randomly chosen people was born in May?*

We obtain problem 1' by just replacing the phrase "the same month" with "May" in problem 1, i.e., we specify the month. Problem 1' has been confusing for many who attempted to solve it. The misunderstanding arose as some would interpret it as equivalent to problem 2. In contrast, to others who would interpret problem 1' as equivalent to problem 3.

**Solutions and analysis of problems**

For simplicity in all these problems, we assume that it is equally likely to be born in any of the 12 months, although different months can have different numbers of days.

1. *What is the probability that two randomly chosen people were not born in the same month?*

   Without loss of generality, let's assume that the first person was born in one of the months of the year. The probability of that event is 1. The probability that the second person was born in a different month from the first person is 11/12 since there are 12 months, and 11 of them satisfy the condition. The probability that the two were born not in the same month will be $1 \cdot (11/12) = 11/12 = 0.91(6)$ since we have independent events.
   Problem 1 could also be formulated differently: *What is the probability that two randomly chosen people were born in different months?* We stick to the first formation as it relates better to the rest of the problems we discuss here.

2. *What is the probability that two randomly chosen people were not both born in May?*

   In this case, we can first find the probability that both were born in May. The probability that the first person was born in May is 1/12 since there are 12 months, and only one of them (May) satisfies the condition. The probability that the second person was born in May is again 1/12. The probability that both were born in May will be $(1/12) \cdot (1/12) = (1/12)^2$ since we have independent events. Hence, the probability that two randomly chosen people were not both born in May will be $1 - (1/12)^2 = 143/144 = 0.9930(5)$.

Some students suggested the following "solution" for problem 2: assume that the first person was born in one of the months. The probability of that event is 1. The second person can be born in any month except May, the probability of that is 11/12 since there are 12 months, and 11 of them (all except May) satisfy the condition. Hence, the probability that the two were born not in May will be 1· (11/12) = 11/12 = 0.91(6) since we have independent events. Notice that there is an error in this approach. Namely, in the first step, where we assumed that the first person was born on one of the months, that person could have been born in Jun, for example, then the second person could have been born in May without violating the condition of the problem however in the second step we imposed that the second person can be born in any month except May.

To solve problem 2 using a different approach, we can calculate the probability that the first person was born in any month except May, which will be 11/12. Then, we calculate the probability that the second person was born in any month except May, which will also be 11/12. The probability that neither was born in May will be $(11/12)^2$. Next, we need to add to this the probability that one of the persons was born in May and the other one was not: this probability can be calculated using the binomial distribution (calculating the probability of one success (a person born in May) out of two trials (since we have two persons)) $(2!/(1!·1!))·(1/12)·(11/12)$ = 2·(1/12)·(11/12). Hence, the probability that the two were born not in May will be $(11/12)^2$ + 2·(1/12)·(11/12) = 143/144 = 0.9930(5). If we missed the binomial coefficient and added (1/12)·(11/12), we would get $(11/12)^2$ + (1/12)·(11/12) = 11/12 = 0.91(6), which was the answer for problem 1 and the incorrect answer for problem 2 that we discussed above.

3. *What is the probability that neither of the two randomly chosen people was born in May?*

   Here, we first calculate the probability that the first person was born in any month except May. This probability will be 11/12. Similarly, the probability that the second person was born in any month except May will also be 11/12. Hence, the probability that neither of them was born in May will be (11/12) · (11/12) = $(11/12)^2$ = 121/144 = 0.8402(7) since we have independent events.
   Once again, notice the difference between problem 2 and problem 3. In the case of problem 2, one of the two persons could have been born in May, but in that case, the other one must have been born in a different month, while in the case of problem 3, neither of them could have been born in May.

Notice that if we replace the word "May" with "the same month" in problems 2 and 3, we will have:
*What is the probability that two randomly chosen people were not both born in the same month May?*
and
*What is the probability that neither of the two randomly chosen people was born in the same month May?*
Both of these problems have the same solution as problem 1, resulting in 11/12 = 0.91(6). This is why, unlike problem 1', problem 1 was not confusing.

**Computational verification**

We can test the answers to some probability theory problems by correctly sampling and repeating many numerical experiments. Thus, students can gain a different perspective on probability theory problems and learn or improve their coding skills. Below, we provide a simple Python code for numerically verifying the answers to the abovementioned problems.

```python
import random
import math

p1 = 0;
p2 = 0;
p3 = 0;
m = list(range(1, 13));
n = 10**7;

for i in range(n):
    a = random.choices(m, k = 2)
    if a[0] != a[1]:
        p1 = p1 + 1
    if a[0] != 5 or a[1] != 5:
        p2 = p2 + 1
    if a[0] != 5 and a[1] != 5:
        p3 = p3 + 1

print ("1. What is the probability that two randomly chosen people were not born in the same month?")
print(*[p1/n, "±", math.sqrt((p1/n)*((1-p1/n)/n))])

print("2. What is the probability that two randomly chosen people were not both born in May?")
print(*[p2/n, "±", math.sqrt((p2/n)*((1-p2/n)/n))])

print("3. What is the probability that neither of the two randomly chosen people was born in May?")
print(*[p3/n, "±", math.sqrt((p3/n)*((1-p3/n)/n))])
```

✓ 23.5s

```
1. What is the probability that two randomly chosen people were not born in the same month?
0.9166901 ± 8.738956491594981e-05
2. What is the probability that two randomly chosen people were not both born in May?
0.9930496 ± 2.627183271079508e-05
3. What is the probability that neither of the two randomly chosen people was born in May?
0.8403846 ± 0.00011581810052096347
```

**Code.** *This simple Python code provides a numerical estimation for the answers to the problems discussed in this paper. We do 10 million numerical experiments here. Inside the "for" loop (line 10), we choose with replacement, uniformly randomly two natural numbers between 1 and 12 (these are months when two randomly chosen people were born) (line 11). In the first "if" statement in line 12, we check whether these two numbers are not equal, if so, it counts for problem 1 (p1). Then, in the "if" statement in line 14, we check if the first number or the second number is not equal to 5 (since May is the fifth month of the year), if so, it counts for problem 2 (p2). Then, in the "if" statement in line 16, we check if the first number and the second number are not equal to 5 (this could have also been written as a[0] != a[1] != 5, more similar to the "if" statement for problem 1 in line 12) if so it counts for problem 3 (p3). In summary, the condition of the "if" statement in line 12 corresponds to the condition of problem 1, the condition of the "if" statement in line 14 corresponds to the condition of problem 2, and the condition of the "if" statement in line 16 corresponds to the condition of problem 3. Besides estimating the probability, we also show an error (due to the finite number of experiments) calculated using the Bernoulli distribution's standard deviation.*

**Chatbots responses**

Large language models recently became very capable of performing various cognitive tasks, including solving mathematics problems (6, 7). We tested two of the popular chatbots (as of December 1, 2024), GPT-4o (4) and Claude 3.5 Sonnet (5), by giving them the problems discussed in this work. Since there is some variability in the responses generated by chatbots, we iterated each problem ten times. Namely, the problems were entered consecutively into the same chat. The procedure was as follows: problem 1 was entered and iterated ten times, then problem 2 was entered and iterated ten times, then problem 3 was entered and iterated ten times, and finally, problem 1' was entered and iterated ten times. These tests were done on December 1, 2024.

Both chatbots provided correct solutions for problems 1, 2, and 3 in all ten iterations. The situation was more interesting with problem 1'. GPT-4o in the eight cases provided answer 143/144 (an example below on the left), and in two cases, it provided answer 121/144 (an example below on the right).

---

What is the probability that two randomly chosen people were not born in May?

To calculate the probability that two randomly chosen people were **not both born in May**, we can use a similar approach:

**Step 1: Calculate the probability of the complementary event**

The complementary event is that **at least one of them is born in May**.

The probability that both people are **born in May** is:

$$P(\text{both born in May}) = \frac{1}{12} \times \frac{1}{12} = \frac{1}{144}$$

**Step 2: Subtract from 1 to find the desired probability**

The probability that the two people are **not both born in May** is the complement:

$$P(\text{not both born in May}) = 1 - P(\text{both born in May}) = 1 - \frac{1}{144} = \frac{143}{144}$$

**Final Answer:**

The probability that two randomly chosen people were **not both born in May** is:

$$\frac{143}{144} \approx 0.9931 \text{ or about } 99.31\%.$$

4/10

---

What is the probability that two randomly chosen people were not born in May?

To find the probability that **two randomly chosen people were not born in May**, at least one of them must be **not born in May**.

**Step 1: Probability that a person is not born in May**

The probability that a single person is not born in May is:

$$P(\text{not born in May}) = 1 - P(\text{born in May}) = 1 - \frac{1}{12} = \frac{11}{12}.$$

**Step 2: Probability that both people are not born in May**

If the events are independent, the probability that both people are not born in May is:

$$P(\text{both not born in May}) = P(\text{not born in May}) \times P(\text{not born in May}) = \frac{11}{12} \times \frac{11}{12} = \frac{121}{144}.$$

**Final Answer:**

The probability that two randomly chosen people were **not born in May** is:

$$\frac{121}{144} \approx 0.8403 \text{ or about } 84.03\%.$$

8/10

The responses of Claude 3.5 Sonnet to problem 1' were more diverse. In two cases, it was more inclined toward the 143/144 answer (an example below on the left). In three cases, it was more inclined toward the 121/144 answer (an example below on the top right). In five cases, it asked for further clarification, stating that both answers are possible depending on what exactly is being asked (an example below on the bottom right).

**Discussion and conclusions**

In this work, we formulated four related problems from probability theory. As we have seen, in some cases, a slight change of wording can completely change the problem. In other cases, changing the problem's wording does not affect the condition and the solution. These problems prove to be useful for students since they emphasize the importance of understanding the actual statement and the requirement of the problem before going to the next step of coming up with the solution plan. It will also be interesting to study how native English-speaking undergraduate students will solve these problems.

A numerical check of the answer to some probability theory problems with widely used programming languages such as Python is helpful for students for not only comparing the answers but also having a different view on the problems, improving coding skills, and learning correct sampling and numerical methods. Given the growing availability of computational tools and enhanced ease of use of programming languages, we advocate for integrating numerical verification using popular programming languages into mathematics education.

Since late 2022, chatbots have become a popular resource for students seeking solutions to mathematics problems. In just two years, the capabilities of chatbots increased significantly, yet the mechanisms of the generated responses and their validity are unclear to the broader public. It would be helpful if chatbots provided confidence levels for their responses. Problems and solutions discussed in this work are recommended to be used in the problem-solving sessions of mathematics and probability theory classes for undergraduate and high school students.

## Acknowledgments

We thank Rafik Santrosyan (AUA) and Mica Hilson (AUA) for the helpful discussions. We thank the students (Silvi Avayan, Elen Sargsyan, Artur Asatryan, Ararat Yenokyan, Suren Mailyan, Arina Pambukyan, Silvi Harutyunyan, Knarik Harutyunyan, Nareg Iskandar Baba, Eduard Gasparyan, Anzhela Margaryan, Sona Martirosyan, Donara Sardaryan, Tigran Sayadyan, Anahit Arzumanyan, Davit Harutyunyan, Hayk Nazaryan) who were taking "Mathematical Thinking" course at AUA and participated in solving earlier versions of problems discussed in this work. We are thankful to the 200 AUA ChangeMakers.

## References

1. J. Rosenhouse, *The Monty Hall problem: the remarkable story of math's most contentious brainteaser* (Oxford University Press, 2009).

2. M. Gardner, *The 2nd Scientific American book of mathematical puzzles & diversions*, University of Chicago Press ed (University of Chicago Press, 1987).

3. A. Zuboff, One self: The logic of experience. *Inquiry* **33**, 39–68 (1990).

4. OpenAI, GPT-4o System Card. [Preprint] (2024). Available at: http://arxiv.org/abs/2410.21276.

5. Introducing Claude 3.5 Sonnet \ Anthropic. Available at: https://www.anthropic.com/news/claude-3-5-sonnet.

6. S. Bubeck, *et al.*, Sparks of Artificial General Intelligence: Early experiments with GPT-4. [Preprint] (2023). Available at: http://arxiv.org/abs/2303.12712.

7. T. H. Trinh, Y. Wu, Q. V. Le, H. He, T. Luong, Solving olympiad geometry without human demonstrations. *Nature* **625**, 476–482 (2024).